# Dual-branch residual network for lung nodule segmentation

Haichao Cao[1], Hong Liu[1*], Enmin Song[1], Chih-Cheng Hung[2], Guangzhi Ma[1], Xiangyang Xu[1], Renchao Jin[1] and Jianguo Lu[1].

[1]Huazhong University of Science and Technology, School of Computer Science & Technology, Wuhan, 430074, China
[2]Kennesaw State University, the Laboratory for Machine Vision and Security Research, 1000 Chastain Rd., Kennesaw, USA, GA 30144

*E-mails: hl.cbib@gmail.com

## Abstract

An accurate segmentation of lung nodules in computed tomography (CT) images is critical to lung cancer analysis and diagnosis. However, due to the variety of lung nodules and the similarity of visual characteristics between nodules and their surroundings, a robust segmentation of nodules becomes a challenging problem. In this study, we propose the Dual-branch Residual Network (DB-ResNet) which is a data-driven model. Our approach integrates two new schemes to improve the generalization capability of the model: 1) the proposed model can simultaneously capture multi-view and multi-scale features of different nodules in CT images; 2) we combine the features of the intensity and the convolution neural networks (CNN). We propose a pooling method, called the central intensity-pooling layer (CIP), to extract the intensity features of the center voxel of the block, and then use the CNN to obtain the convolutional features of the center voxel of the block. In addition, we designed a weighted sampling strategy based on the boundary of nodules for the selection of those voxels using the weighting score, to increase the accuracy of the model. The proposed method has been extensively evaluated on the LIDC dataset containing 986 nodules. Experimental results show that the DB-ResNet achieves superior segmentation performance with an average dice score of 82.74% on the dataset. Moreover, we compared our results with those of four radiologists on the same dataset. The comparison showed that our average dice score was 0.49% higher than that of human experts. This proves that our proposed method is as good as the experienced radiologist.

# 1. Introduction

Lung cancer is a relatively common and deadly cancer with a five-year survival rate of only 18% [1]. The use of computed tomography (CT) images for treatment, monitoring, and analysis is an important strategy for early lung cancer diagnosis and survival time improvement [2]. With this technique, the accurate segmentation of lung nodules is important because it can directly affect the subsequent analysis results [3]. Due to the fact that the number of CT images is increasing, the development of a robust automatic segmentation model has important clinical significance for avoiding tedious manual treatment and reducing the diagnostic difference among doctors [4].

Due to the heterogeneity of lung nodules on CT images (as shown in Fig. 1), it has been difficult to obtain an accurate segmentation performance [5–7]. Specifically, the similarity of visual characteristics between nodules and their surroundings causes the difficulty for the segmentation. In particular, the juxtapleural nodules (Fig. 1(b)), because its intensity is very similar to the lung wall, makes it difficult to segment such nodules using conventional methods. A similar situation is the ground-glass opacity (GGO, Fig. 1(e)) nodules, which, due to their low contrast to the surrounding background, result in simple threshold- and morphological-based methods that cannot handle such nodules. In addition, for calcific nodules (Fig. 1(d)), because of the high contrast with surrounding pixels, a simple threshold segmentation method (for example, OTSU algorithm) can segment such nodules well, but such methods cannot be applied to both the juxtapleural nodules and the GGO nodules. It is a challenge to adapt to these three types of nodules at the same time. Finally, for the cavitary nodules with a black hole (Fig. 1(c), since the intensity difference of each part is large, it is also a challenge to accurately segment such nodules. It should be pointed out that there are some nodules with small diameters in the lungs as shown in Fig. 1(f). These nodules are very similar to the intensity of the surrounding noise, which makes these nodules more difficult to be distinguished.

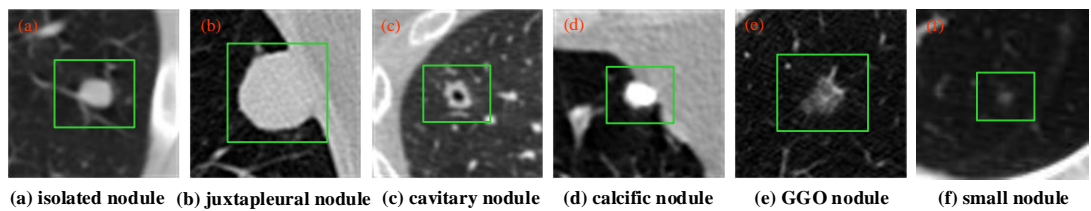

(a) isolated nodule  (b) juxtapleural nodule  (c) cavitary nodule  (d) calcific nodule  (e) GGO nodule  (f) small nodule

Fig. 1. Example image of a heterogeneous lung nodule in CT image. Note that (e) GGO in sub-figure (d) represents a ground-glass opacity nodule, and sub-figure (f) is a small nodule having a diameter of less than 4.4 mm.

The intensity-based method using morphological operations [8,9] and region growing [5,10] were used for lung nodule segmentation. In addition, energy-optimized methods such as graph cut [11] and level set [7] have also been used. However, both approaches are not robust for segmenting juxtapleural nodules and small nodules with a diameter of less than 6 mm. For example, in a morphology-based method, the size of the morphological template is difficult to be adaptive for the nodules of various diameters [5]. Some effective measures are semi-automatic interaction methods that require user intervention [12] and shape-constrained methods based on specific rules [7,13]. However, this method may fail for irregular nodules due to the violation of shape assumption. The limitations of segmentation directly using raw gray values indicate that a robust method for segmentation of lung nodules is urgently needed.

In recent years, in the field of medical image segmentation, convolutional neural networks (CNN) have achieved good performance [14–16]. However, for various types of lung nodules as shown in Fig. 1, the applicability of CNN-based methods is not fully explored yet.

In order to adapt to the heterogeneity of lung nodules, we followed the voxel classification scheme and proposed a Dual-Branch Residual Network (DB-ResNet), which is suitable for various types of lung nodule segmentation. In general, our technical contributions in this work have the following four aspects.

(1) For small nodules and juxtapleural nodules, the proposed DB-ResNet model can achieve attractive segmentation performance (Fig. 1).

(2) A Dual-branch CNN architecture based on ResNet is proposed in which the extracted multi-view and multi-scale features are used to classify each voxel (Section 3.1.2). In this architecture, multi-view branches are used to model the upper, middle, and lower slices while the multi-scale branches are used to model the three different scales of the middle slice (Fig. 2).

(3) We proposed a central intensity-pooling layer that preserves the intensity features centered on the target voxel rather than the intensity information at the boundary. We incorporate the traditional intensity features into the CNN architecture to achieve a performance improvement in the nodule segmentation model (Section 3.1.3).

(4) The weighted sampling strategy [17] was improved to handle the unbalanced training labels to achieve efficient model training. In this improved sampling strategy, the small nodules can be adequately sampled based on the number of voxels located at the boundary of the nodules (Section 3.2).

# 2. Related Work

In recent years, many methods for segmentation of lung nodules have been proposed, such as morphological based methods, region growing based methods, energy based optimization methods, and machine learning based methods. Below we will further describe these four types of methods.

In the morphology method, in order to remove the nodule-attached vessels, morphological operations were applied and lung nodules were then isolated according to the selection of connected regions [18,19]. Further, in order to better separate the lung wall from the juxtapleural nodules, a morphological operation combining the shape hypothesis was introduced, replacing the fixed size morphological template [20,21]. In addition, a 2-D rolling ball filter [9] has also been proposed to process juxtapleural nodules. In general, the segmentation of nodules is very challenging by using morphological operations [8].

In the region growing method, we first need to specify the seed point for the region growing, and then iterate until the termination condition is met. Such methods are only well adapted to isolate calcified nodules, but are not able to segment the nodule similar to the juxtapleural nodules. In order to alleviate this problem, Dehmeshki et al. proposed a new region growing method based on intensity information, distance, fuzzy connectivity and peripheral contrast [10]. Although Dehmeshki et al. introduced a variety of rules, they still do not adapt well to irregularly shaped nodules because they have almost no rules to follow. There are similar problems, as well as the segmentation method of lung nodules based on convexity model and morphological operation proposed by Kubota et al. [5].

In the energy optimization method, people usually convert the segmentation task into an energy minimization task to process. For example, in [22–25], the author uses a level set function to characterize the image, and when the segmented contour matches the nodule boundary, the energy function reaches a minimum. A similar approach is the lung nodules segmentation method based on shape prior hypotheses and level sets, proposed by Farag et al. [7]. In addition, the graph cut method that converts the lung nodule segmentation task into the maximum flow problem is also used [11,26,27]. However, these methods are not well adapted to the GGO nodules and the juxtapleural nodules.

In the machine learning method, in order to segment the target, related features need to be designed and extracted for subsequent voxel classification [28–31]. For example, Lu et al. designed a set of features with translational and rotational invariance that played a positive role in classification [32]. Wu et al. proposed a method for segmentation of lung nodules based on conditional random fields, which extracted the texture and shape features of nodules [33]. Hu et al. segmented the lungs and then performed vascular feature extraction based on the Hessian matrix to obtain the mask of the lung blood vessels. The blood vessels are then removed from the lung mask and the artificial neural networks are used for the classification [34].

CNN is conceptually similar to previous machine learning-based methods, which is also the solution of transforming the segmentation task into the voxel classification problem. For example, Ciresan et al. used CNN to classify voxels in microscopic images to segment neuronal cell membranes [35]. Similarly, Zhang et al. apply depth CNN to classify voxels in MR images to obtain the mask of infant brain tissue [16]. In addition, the CNN models using multiple views [36], multiple branches [37], or a combination of both [38] have also been proposed for segmentation of lung nodules. Wang et al. proposed a semi-automatic central focused convolutional neural network [17] for voxels classification, however, the model is not ideal for small nodules. On the other hand, full convolutional neural network (FCN) [39] is another approach of image segmentation. For example, the 2D UNet network architecture proposed by Ronneberger et al. [40] and the 3D UNet network architecture proposed by Çiçek et al. [41] are a kind of segmentation method that can better adapt to medical images.

# 3. Our Proposed Method

We will describe our proposed method in detail below. The method is divided into three components: 1) the model architecture, 2) the sampling strategy, and 3) the post-processing approach.

## 3.1. The Model Architecture

The proposed DB-ResNet model utilizes three longitudinal views (three contiguous slices) and three transversal scales to segment the lung nodules. Given a voxel in a slice of the CT images, we extract multiple views from the slices centered in the current voxel and multiple different size of patches as multiple scales. In this study, we limit the multiple views and scales to three views and scales. Three views are taken from the previous, current and next slices. The multiple views and scales will be used as input, and then output the probability that this voxel belongs to the nodule. Fig. 2 shows the proposed architecture of DB-ResNet. Table 1 shows the corresponding network parameters.

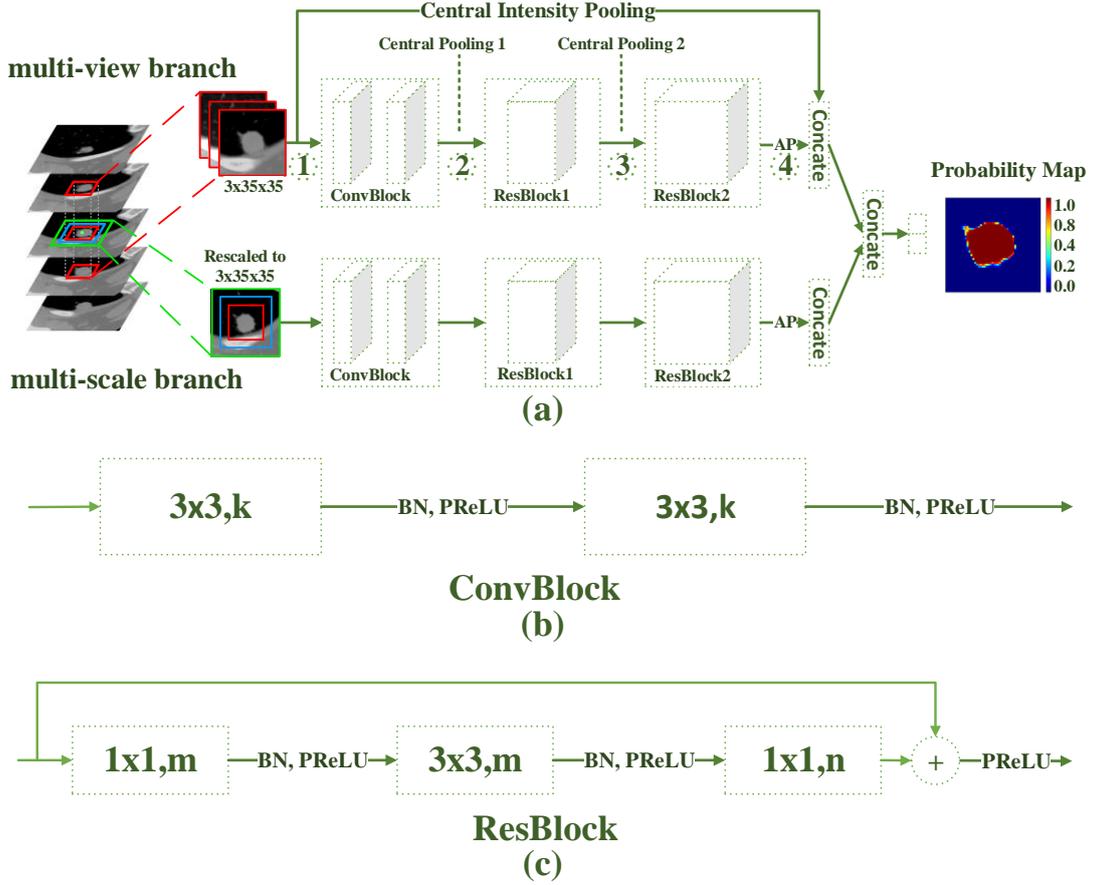

Fig. 2. (a) Illustration of the proposed DB-ResNet architecture where AP and Concate represent the Average Pooling operation and Concatenate operation, respectively. The symbol ⦂N⦂ indicates where the Central Intensity-Pooling can be placed, (b) the diagram of the convolution block (ConvBlock), and (c) the diagram of the residual block (ResBlock). The parameters k, m and n indicate the number of channels.

Table 1. Network parameters of the DB-ResNet. Building blocks are shown in brackets with the numbers of blocks stacked. Downsampling is performed using Central Pooling before the first layer of ResBlock1_x and ResBlock2_x.

| Layer name | Output size | 32-Layer | 83-Layer | 134-Layer |
|---|---|---|---|---|
| ConvBlock_x | 35×35 | $\begin{bmatrix} 3{\times}3,\ 36 \end{bmatrix}{\times}2$ | $\begin{bmatrix} 3{\times}3,\ 36 \end{bmatrix}{\times}2$ | $\begin{bmatrix} 3{\times}3,\ 36 \end{bmatrix}{\times}2$ |
| ResBlock1_x | 17×17 | $\begin{bmatrix} 1{\times}1,\ 128 \\ 3{\times}3,\ 128 \\ 1{\times}1,\ 512 \end{bmatrix}{\times}4$ | $\begin{bmatrix} 1{\times}1,\ 128 \\ 3{\times}3,\ 128 \\ 1{\times}1,\ 512 \end{bmatrix}{\times}4$ | $\begin{bmatrix} 1{\times}1,\ 128 \\ 3{\times}3,\ 128 \\ 1{\times}1,\ 512 \end{bmatrix}{\times}8$ |
| ResBlock2_x | 8×8 | $\begin{bmatrix} 1{\times}1,\ 256 \\ 3{\times}3,\ 256 \\ 1{\times}1,\ 1024 \end{bmatrix}{\times}6$ | $\begin{bmatrix} 1{\times}1,\ 256 \\ 3{\times}3,\ 256 \\ 1{\times}1,\ 1024 \end{bmatrix}{\times}23$ | $\begin{bmatrix} 1{\times}1,\ 256 \\ 3{\times}3,\ 256 \\ 1{\times}1,\ 1024 \end{bmatrix}{\times}36$ |
| | 1×1 | Average Pooling, Concatenate, 2-d fc, softmax | | |
| Params | | $0.68{\times}10^7$ | $2.3{\times}10^7$ | $3.7{\times}10^7$ |

### 3.1.1. Network structure

The network contains two deep branches that share the same structure, but the inputs used for training are different. Each branch of the proposed network architecture contains 32 convolutional layers, two central pooling layers [17], one central intensity-pooling (CIP) layer (see Section 3.1.3 for a detailed description) and one shared fully-connected layer. The 32 convolution layers in the CNN are divided into three categories: the first is a ConvBlock consisting of two convolutional layers, the second is a ResBlock cluster consisting of four residual blocks [42], and the last is a ResBlock cluster consisting of six residual blocks. In order to speed up the training process, each convolutional layer is batch-normalized to normalize the corresponding output [43]. After each convolution layer, a parametric rectified linear unit (PReLU) is used as a nonlinear activation function [44].

We use the average pooling (Fig. 2, AP) for the output of the last convolutional layer of each branch. The output of the average pooling is then concatenated with the output of the CIP layer to fuse the depth features produced by the convolution layer and the intensity features generated by the CIP layer. At the end of the model layers, the features generated by the two CNN branches are combined with the concatenation, and the concatenated results are then connected to a fully-connected layer to capture the correlation of the features generated by the two CNN branches.

The goal of network training is to maximize the probability of the correct class for each voxel. We achieve this by minimizing the cross-entropy loss of each training sample. For a given input patch belonging to {0, 1}, assuming that $y_n$ is a true label, then the loss function is defined as shown in Equation (1):

$$L = -\frac{1}{N}\sum_{n=1}^{N}[y_n log(y_n{}') + (1 - y_n)log(1\text{-}y_n{}')] \tag{1}$$

Where $y_n{}'$ represents the prediction probability of the DB-ResNet, and N is the number of samples.

In the experiment, we used the Stochastic Gradient Descent (SGD)) algorithm [45] as a model update method. The SGD optimizer has several parameter settings: the initial learning rate is 0.001, and then the learning rate is decreased by ten percent in every five epochs. In addition, the momentum setting is 0.9. However, due to the limitation of GPU memory, only a batch size of 32 samples are used. In order to avoid overfitting during the training process, we adopted the early stopping training strategy [46].

### 3.1.2. Dual-Branch Architecture

The proposed dual-branch residual network (DB-ResNet) architecture aims to capture both multi-view features in multiple slices and multi-scale features in the current slice.

The input size of the multi-view branch is a $3 \times 35 \times 35$ 3D data patch. Specifically, for a voxel, we extend the current, previous, and subsequent slices centered on this voxel to extract training patches (see Fig. 2, Multi-view Branch). This three-slice patch extracted are treated as three channel images and fed to the multi-view CNN branch.

Simultaneously, we have introduced a multi-scale branch trying to focus on learning features from the current slice because of their high resolution in all CT scans. The purpose of designing multi-scale branches is to model the relationship among three-scale patches through the feature extraction layer. Firstly, three image patches with a size of 65×65, 50×50 and 35×35, respectively, were extracted on the target voxels from three slices. They are then rescaled to the same size of 35×35 using a third-order spline interpolation and forming three-channel patches as input to a multi-scale CNN branch (see Fig. 2, multi-scale branch).

In addition, in order to further improve the segmentation performance, we also integrate the residual learning structure into the network. Moreover, we use the bottleneck structure where the head and end are 1x1 convolutions (to reduce and restore dimensions) and the middle is a 3x3 convolution, replacing the original residual learning structure, which can reduce network parameters and increase network depth [42].

### 3.1.3. The Central Intensity-Pooling

The conventional segmentation method usually utilizes the intensity information of the target. For the segmentation of nodules, we can also use the same information. In particular, for isolated nodules and calcified nodules, the intensity information is useful due to a large contrast between the nodules and the surrounding background. Therefore, we designed a pooling layer that calculates either the center position of the feature map or its surrounding intensity information.

Fig. 3 shows the central intensity-pooling process for three different pooling kernel sizes. Among them, the yellow mark corresponds to a pooling process with a pooing kernel size of 1x1, and the result is the intensity value of the pixel in the center of the input image. The blue mark corresponds to the pooling process with a pooling kernel size of 3x3, and the result is the average intensity value of the surrounding 3x3 region centered on the center pixel of the input image. The corresponding pooling process for the red mark is similar to the blue mark. In practice, we designed two different sizes of the pooling kernel. One is a smaller local pooling kernel that can obtain local intensity information at the center of the image; the other is a larger global pooling kernel that can obtain richer contextual information. Since we predict the category of the center voxel of the patch, the proposed central intensity-pooling helps to extract the intensity features at the center of the patch.

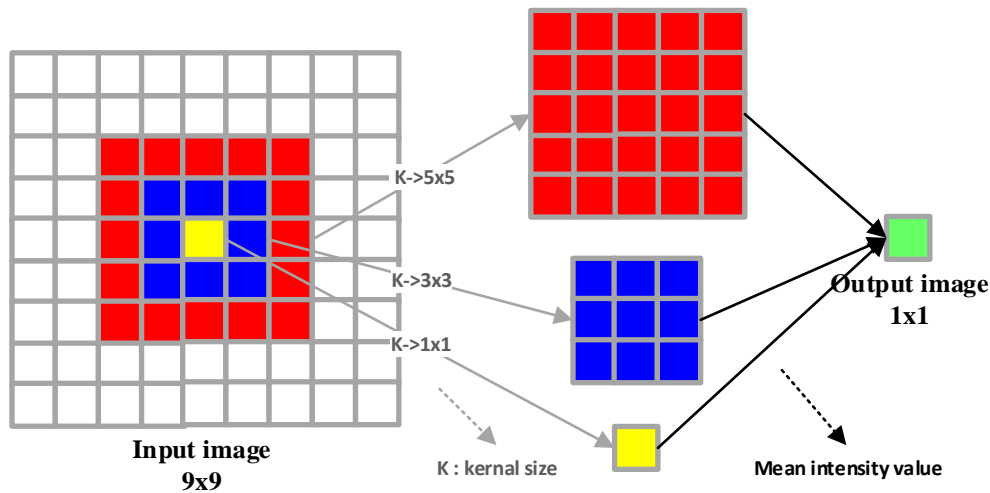

Fig. 3. A central intensity-pooling process: this shows the processing of three different pooling kernel sizes, corresponding to the red, blue, and yellow, and the sizes are 5x5, 3x3, and 1x1, respectively.

This central intensity-pooling consists of two parameters: 1) the size of the different pooling kernel and 2) the number of pooling kernels for each type. As mentioned earlier, in this study, we have introduced two different sizes of the pooling kernel, and the number of pooling kernel for each type has only one. These two types correspond to the local pooling kernel and the global pooling kernel. In our experiment, the size of the local pooling kernel is 1x1, and the size of the global pooling kernel is 3x3.

## 3.2. The weighted sampling strategy

Since our approach focuses on automatically learning advanced semantic features from images, a large number of voxel patches are needed as training samples to improve the accuracy of the model. However, in a CT slice, the ratio of nodule to non-nodular voxel is generally 1:370 ($\pi r^2 : S - \pi r^2$, where r=15 is the maximum radius of the nodule and S=$512^2$ is the area of each slice), which is a highly data imbalance problem. If a traditional random sampling is used, this will lead to a trained model that is biased towards the non-nodal classes. Therefore, in order to avoid this problem, we use a weighted sampling strategy [17]. However, our experimental results have found that this weighted sampling strategy has poor sampling results for small nodules with a diameter of less than 6 mm.

To elaborate further on the above issue, we assume that the nodules in each slice are circular, and the nodule diameter of the $k^{th}$ slice is R. Then, the total number of nodule voxels and nodule voxels at the boundary in the $k^{th}$ slice can be approximated as $\pi R^2 / 4$ and $\pi R$, respectively. According to the original weighted sampling method, only 40% of the total number of nodal voxels is sampled. If R is less than 10, the number of sampling points of the nodule class will be smaller than the number of the voxels sampled at the boundary. In our experiment, we found that if a nodule is less than 6mm in diameter, it will have almost half of the voxels at their boundary are not sampled.

In order to solve the problem of insufficient number of samples for small nodules, we set the number of nodule samples to twice the number of voxels at the nodule boundary. Simultaneously, we also ensure that the number of non-nodule samples is the same as the number of nodule samples. It should be noted that for small nodules that are less than 6 mm in diameter, the total number of nodule voxels might be less than twice the number of boundary voxels. In this case, we will take all the voxels of such small nodules to improve the generalization capability of the model for such small nodules. Experimental results have shown that this improved sampling strategy increases the average dice score from 78.89% to 80.30%. The detailed results are given in Table 3 of Section 4.4.

## 3.3. Post-processing

Since the method proposed in this paper is a semi-automatic segmentation model, it is necessary to give the volume of interest (VOI) where the nodule is located before segmentation. However, since the nodules are usually distributed over multiple CT slices, it is tedious to manually specify the region of interest (ROI) in which the nodules are located, layer-by-layer. To facilitate the doctor's operation, we performed the following post-processing operation, that is, it is only necessary to manually designate a bounding box called a starting slice on one CT slice.

Then repeat applying the same bounding box to the previous and next slices until the following experimental conditions are satisfied: The nodule intersection area of the current slice and the previous slice is less than 30% of the nodule area in the previous slice.

To remove the noisy voxels, we made a simple connected region selection as follows: 1) When noise appears in the starting slice, we select the isolated region closest to the center of the bounding box, and 2) when noise occurs in other slices, we choose the connected region where the overlap O=V(Gt∩Seg)/V(Gt∪Seg) (will be explained in detail in Section 4.2) of the current slice and the previous slice nodule mask is the largest.

# 4. Data and Experiments

We give the information of the dataset and experiments in detail in this section. The evaluation criteria and the ablation study of the proposed method are described below.

## 4.1. Data

We used public datasets from the Lung Image Database Consortium and Image Database Resource Initiative (LIDC) in our experiments and for comparison [47–49]. In this study, we studied 986 nodule samples annotated by four radiologists. Due to the differences in labeling between the four radiologists, the 50% consistency criterion [5] was used to generate the ground-truth boundary.

We randomly partitioned 986 nodules into three subsets for training, validation, and testing with the number of nodules contained in each subset was 387, 55, and 544, respectively. As shown in Table 2, the clinical characteristics of the three subsets have a similar statistical distribution.

Table 2. The data distribution of the LIDC dataset training, validation and testing sets. Among them, the values are displayed in the format of mean ± standard deviation.

| Characteristics | Training set （n=387) | Validation Set (n=55) | Test set (n=544) |
|---|---|---|---|
| Diameter(mm) | 8.34±4.73 | 8.17±4.61 | 7.90±4.14 |
| Sphericity | 3.80±0.58 | 3.84±0.62 | 3.85±0.58 |
| Margin | 4.07±0.73 | 4.06±0.81 | 4.11±0.78 |
| Spiculation | 1.61±0.78 | 1.54±0.69 | 1.57±0.74 |
| Texture | 4.56±0.83 | 4.45±0.98 | 4.57±0.80 |
| Calcification | 5.65±0.80 | 5.68±0.77 | 5.67±0.80 |
| Internal structure | 1.01±0.16 | 1.03±0.20 | 1.01±0.08 |
| Lobulation | 1.74±0.72 | 1.75±0.74 | 1.69±0.71 |
| Subtlety | 4.00±0.78 | 3.89±0.74 | 3.95±0.75 |
| Malignancy | 2.95±0.91 | 2.87±0.77 | 2.91±0.91 |

Note: The range of all characteristic values except diameter, internal structure and calcification is 1-5, wherein the internal structure and calcification range from 1 to 4, 1 to 6, respectively. Margin indicates the clarity of the nodule edge. Lobulation and spiculation indicate the number of these shapes. Texture is a statistic of the distribution properties of the local gray information of nodules.

Internal structural represents the internal composition of the nodule. Malignancy, calcification, and Sphericity indicate the possibility that the nodule is such a feature. Subtlety describes the contrast of the nodule region and its surrounding region. There were no significant statistical differences in the characteristics of the three subsets.

## 4.2. Evaluation criteria

To evaluate the segmentation results of the DB-ResNet model, we used the average surface distance (ASD) and dice similarity coefficient (DSC) as the primary evaluation criteria. DSC is a metric that is widely used to measure the overlap between two segmentation results [15,37]. Moreover, in order to ensure the robustness of the evaluation, we also use the true prediction value (PPV) and sensitivity (SEN) as auxiliary evaluation parameters. The entire definition is shown in formulae (2)-(5).

$$DSC = \frac{2 \times V(Gt \cap Seg)}{V(Gt) + V(Seg)} \qquad (2)$$

$$ASD = \frac{1}{2}(mean_{i \in Gt} min_{j \in Seg} d(i, j) + mean_{i \in Seg} min_{j \in Gt} d(i, j)) \qquad (3)$$

$$SEN = \frac{V(Gt \cap Seg)}{V(Gt)} \qquad (4)$$

$$PPV = \frac{V(Gt \cap Seg)}{V(Auto)} \qquad (5)$$

Among them, "Gt" represents the result of expert labeling; "Seg" represents the segmentation result of DB-ResNet model. V represents the volume size calculated in voxel units and d (i, j) represents the Euclidean distance between the voxel i and voxel j measured in millimeters.

## 4.3. The Detail of Implementation

In the experiment, we used a weighted sampling strategy (Section 3.2) to sample 0.47 million voxel patches extracted from the LIDC training set. To avoid overfitting, we used a training strategy for early stopping: if there is no more improvement in performance, it will stop in an extra training with 10 epochs. It has been found through experiments that the DB-ResNet model generally stops around the 16th epoch, so we set the upper limit of the training epoch to 20. Our experiment is based on the Keras deep learning framework and the coding language is Python 3.6. Our experiment was carried out on a server equipped with an Intel Xeon processor and 125GB memory. In the model training, acceleration is performed on the NVIDIA

GTX-1080Ti GPU (11GB video memory), and the DB-ResNet model takes about 31 hours to converge.

## 4.4. Ablation Study

To verify the effectiveness of each component in the DB-ResNet architecture, we designed an ablation experiment based on the CF-CNN network architecture [17]. The relevant experimental results are shown in Table 3.

Table 3. Ablation study on LIDC testing dataset. Note that Scale represents the 50*50 size of the multi-scale branch; BWS represents a weighted sampling strategy based on the boundary points; DB represents dual-branch architecture; ResNet represents the residual network, see Table 1; CIP_N denotes adding a central intensity-pooling layer from the first to the $N^{th}$ position; Post indicates the proposed post-processing operation.

| Method | DSC | ASD | SEN | PPV |
|---|---|---|---|---|
| CF-CNN | $78.55 \pm 12.49$ | $0.27 \pm 0.35$ | $86.01 \pm 15.22$ | $75.79 \pm 14.73$ |
| CF-CNN + Scale | $78.89 \pm 11.67$ | $0.26 \pm 0.29$ | $86.21 \pm 14.66$ | $75.95 \pm 14.41$ |
| CF-CNN + Scale + BWS | $80.30 \pm 11.34$ | $0.26 \pm 0.45$ | $85.40 \pm 13.27$ | $78.69 \pm 14.49$ |
| DB-ResNet32 | $82.37 \pm 10.98$ | $0.22 \pm 0.34$ | $88.36 \pm 13.09$ | $79.58 \pm 13.30$ |
| DB-ResNet83 | $81.33 \pm 11.69$ | $0.24 \pm 0.39$ | $86.94 \pm 14.42$ | $79.33 \pm 14.08$ |
| DB-ResNet134 | $79.56 \pm 11.28$ | $0.25 \pm 0.36$ | $87.92 \pm 13.24$ | $75.35 \pm 14.66$ |
| DB-ResNet32 + CIP_1 | $82.54 \pm 10.20$ | $0.19 \pm 0.21$ | $89.06 \pm 11.79$ | $79.17 \pm 13.31$ |
| DB-ResNet32 + CIP_2 | $82.69 \pm 10.46$ | $0.21 \pm 0.30$ | $88.69 \pm 12.18$ | $79.62 \pm 13.29$ |
| DB-ResNet32 + CIP_3 | $81.67 \pm 10.46$ | $0.21 \pm 0.25$ | $88.93 \pm 12.32$ | $77.94 \pm 13.68$ |
| DB-ResNet32 + CIP_4 | $80.52 \pm 11.45$ | $0.23 \pm 0.37$ | $88.89 \pm 12.89$ | $76.14 \pm 14.97$ |
| DB-ResNet32+ CIP_1 + Post | $82.74 \pm 10.19$ | $0.19 \pm 0.21$ | $89.35 \pm 11.79$ | $79.64 \pm 13.34$ |

（1）Effect of Boundary-based Weighted Sampling (BWS)

In Table 3, CF-CNN + Scale indicates that we added a 50x50 scale to the 2-D branch of CF-CNN and then combined it with two scales of 65x65 and 35x35 to form our multi-scale branch. The DSC obtained by CF-CNN + Scale is 78.89%, which is slightly higher than that of CF-CNN. Then, based on CF-CNN + Scale, a weighted sampling strategy based on the boundary points is applied, and the DSC obtained is 80.30%. Compared to CF-CNN + Scale, its performance has improved by nearly 1.5%, which verifies the effectiveness of the boundary-based weighted sampling strategy.

（2）Effect of Residual Network

In Table 3, based on CF-CNN + Scale + BWS, DB-ResNet32 replaces two convolution blocks with two residual blocks. At this time, the DSC is 82.37%, which is an increased performance of two percent compared to the previous 80.30%. This proves the effectiveness of the residual block. Then, based on the ideas of ResNet101 and ResNet152 [42], we improve the network performance by increasing the depth of the network, but according to the results of the sixth and seventh rows in Table 3, it does not achieve what we expected. This may be due to the excessive complexity of the network, which leads to overfitting of the model.

（3）Effect of Central Intensity-Pooling

Based on DB-ResNet32, we integrated the proposed central intensity-pooling layer into DB-ResNet32. In order to verify the effectiveness of the central intensity -pooling layer, we performed four experiments, corresponding to rows 8-11 in Table 3. By comparing these four rows, we can see that DB-ResNet32 + CIP_1 is the best with an ASD of 0.19, which is an improvement of three percent over DB-ResNet32. For the other three performance indicators, both the DSC and the SEN are increased except the PPV is decreased by 0.41%. For the reason why the performance of DB-ResNet32 + CIP_3 and DB-ResNet32 + CIP_4 decline more obviously, our opinion is that the features used for classification, the proportion of traditional intensity features is increasing, even exceeding the deep convolutional features. This is unreasonable because the deep convolution feature in our network is crucial. Specifically, for DB-ResNet32 + CIP_3, the ratio of intensity features to convolution features is 584:1024, and the ratio for DB-ResNet32 + CIP_4 is 1608:1024.

（4）Effect of Post-processing

Finally, we verified the effectiveness of the proposed post-processing method. By comparing the ninth row and the last row in Table 3, it can be seen that although the performance is not significantly improved, the four performance measures are improved.

# 5. Results and Discussion

We give the overall performance of our method, the robustness of the proposed segmentation model, and the experimental comparison with other methods below.

## 5.1. Overall performance

To better observe the performance of the proposed method in the testing set, we plot the histogram between the DSC value and the number of nodules, based on all samples in the testing set, as shown in Fig. 4. By observing Fig. 4, we can easily conclude that most of the nodules have a DSC value higher than 0.8.

To see if the segmentation results of our proposed method are comparable to those hand-labeled by human experts, we performed a consistency comparison between DB-ResNet and four radiologists, as shown in Table 4. Our results show that the stability of DB-ResNet is slightly weaker than that of four different radiologists. However, the DSC between DB-ResNet and each radiologist is 83.15% on average, which is higher than the average of 82.66% among inter-radiologists.

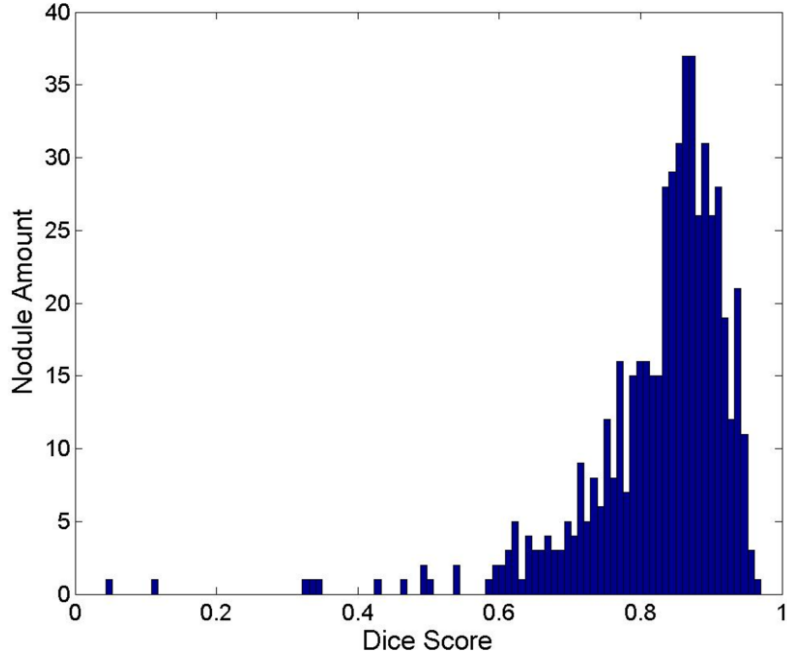

Fig. 4. DSC distributions of the LIDC testing set

Table 4. Mean DSCs (%) of consistency comparison between DB-ResNet and each radiologist, where R1 to R4 represent four radiologists.

|  | R1 | R2 | R3 | R4 | Average |
|---|---|---|---|---|---|
| R1 | – | 82.61 | 82.47 | 82.49 | |
| R2 | 82.61 | – | 83.72 | 82.36 | |
| R3 | 82.47 | 83.72 | – | 82.32 | 82.66 ± 0.48 |
| R4 | 82.49 | 82.36 | 82.32 | – | |
| DB-ResNet | 82.32 | 84.02 | 82.94 | 83.30 | 83.15 ± 0.62 |

## 5.2. Robustness of Segmentation

In order to prove the robustness of the proposed method, we base the nine characteristics corresponding to each nodule as the benchmark, and divide the testing set into different groups according to the characteristic scores of the nodule. Table 5 lists the DSC average values of different nodule groups. As can be seen from Table 5, DB-ResNet can handle all types of nodules with similar performance, which reflects the segmentation robustness of our method.

Further, we have collated the evaluation results of challenging small nodules and attached nodules. The relevant results are shown in Table 6. According to the experimental results in Table 6, it can be seen that the potential robust segmentation of the DB-ResNet is independent of the type of nodules and the size of nodules.

Table 5. The DSC average values on different nodule groups.

| Characteristics | Characteristic scores | | | | | |
|---|---|---|---|---|---|---|
| | 1 | 2 | 3 | 4 | 5 | 6 |
| Malignancy | 78.16 [44] | 81.57 [143] | 83.24 [206] | 84.57 [135] | 83.96 [16] | – |
| Sphericity | – | 78.00 [13] | 81.89 [96] | 82.62 [393] | 87.30 [42] | – |
| Margin | – | 75.61 [33] | 82.50 [60] | 82.66 [281] | 84.35 [170] | – |
| Spiculation | 82.99 [300] | 82.05 [192] | 83.83 [24] | 83.74 [26] | 84.63 [2] | – |
| Texture | 65.18 [7] | 79.67 [22] | 80.53 [8] | 81.69 [117] | 83.59 [390] | – |
| Calcification | – | – | 78.85 [23] | 82.10 [39] | 85.68 [30] | 82.80 [452] |
| Internal structure | 82.82 [541] | 67.89 [3] | – | – | – | – |
| Lobulation | 82.72 [235] | 82.44 [249] | 84.14 [39] | 83.87 [21] | – | – |
| Subtlety | 65.53 [1] | 77.94 [28] | 80.30 [88] | 82.26 [308] | 87.06 [119] | – |

Table 6. In the LIDC testing sets，DSCs and ASDs for nodules attached，non-attached, less than 6mm and more than 6mm in diameter.

| | LIDC testing set | | LIDC testing set | |
|---|---|---|---|---|
| | Attached (n=131) | Non-attached (n=413) | Diameter<6mm (n=241) | Diameter>=6mm (n=303) |
| DSC (%) | 81.79 | 83.04 | 79.97 | 84.94 |
| ASD (mm) | 0.25 | 0.17 | 0.16 | 0.21 |

## 5.3. Experimental Comparison

To illustrate the efficiency of the proposed method, we compared the results with other methods. Two different comparisons are provided: 1) a comparison with various different types of segmentation methods recently proposed and 2) a comparison on the same network architecture with the basic components of the network are different.

Table 7 shows the quantification results for the different types of segmentation methods. The results are in the format of "mean ± standard deviation." In order to ensure the fairness of the comparison, the methods compared with DB-ResNet in Table 7, the conditions of the experiments are consistent with DB-ResNet including boundary-based sampling strategy, central intensity-pooling layer and post-processing methods. According to the experimental results shown in Table 7, the proposed method is superior to the existing segmentation methods.

Table 7. Mean ± standard deviation of the results for various segmentation methods. The best performance is indicated in bold font.

| Network Architecture | DSC (%) | ASD (mm) | SEN (%) | PPV (%) |
|---|---|---|---|---|
| FCN-UNet [40] | 77.84 ± 21.74 | 1.79 ± 7.52 | 77.98 ± 24.52 | 82.52 ± 21.55 |
| CF-CNN [17] | 78.55 ± 12.49 | 0.27 ± 0.35 | 86.01 ± 15.22 | 75.79 ± 14.73 |
| MC-CNN [50] | 77.51 ± 11.40 | 0.29 ± 0.31 | 88.83 ± 12.34 | 71.42 ± 14.78 |
| MV-CNN [51] | 75.89 ± 12.99 | 0.31 ± 0.39 | 87.16 ± 12.91 | 70.81 ± 17.57 |
| MV-DCNN [38] | 77.85 ± 12.94 | 0.33 ± 0.36 | 86.96 ± 15.73 | 77.33 ± 13.26 |
| MCROI-CNN [52] | 77.01 ± 12.93 | 0.30 ± 0.35 | 85.45 ± 15.97 | 73.52 ± 14.62 |
| Cascaded-CNN [37] | 79.83 ± 10.91 | 0.26 ± 0.34 | 86.86 ± 13.35 | 76.14 ± 13.46 |
| **DB-ResNet** | **82.74 ± 10.19** | **0.19 ± 0.21** | **89.35 ± 11.79** | **79.64 ± 13.34** |

Table 8 shows the quantification results of several segmentation methods of the same architecture but with different components. The results are also shown in the format of "mean ± standard deviation". In order to achieve a fair comparison, in Table 8, except for the basic components, the other testing conditions are the same. By comparing the experimental results in rows 2 to 8 in Table 8, we can conclude that the DB-ResNet performs the best.

Table 8. Mean ± standard deviation of quantitative results of segmentation methods using different basic network architectures. The best performance is indicated in the bold font.

| Network Architecture | DSC (%) | ASD (mm) | SEN (%) | PPV (%) |
|---|---|---|---|---|
| DB-VGG [53] | 80.30 ± 11.34 | 0.26 ± 0.45 | 85.40 ± 13.27 | 78.69 ± 14.49 |
| DB-GoogLeNet [54] | 80.61 ± 10.38 | 0.23 ± 0.29 | 86.51 ± 12.76 | 78.03 ± 13.78 |
| DB-Inception-V3 [55] | 81.90 ± 10.61 | 0.22 ± 0.34 | 87.74 ± 13.57 | 79.51 ± 13.53 |
| DB-Inception-V4 [56] | 80.68 ± 12.40 | 0.26 ± 0.45 | 84.67 ± 15.44 | 80.07 ± 14.55 |
| DB-DenseNet [57] | 80.52 ± 11.13 | 0.24 ± 0.32 | 86.44 ± 13.95 | 77.98 ± 13.83 |
| DB-ResDenseNet [58] | 79.08 ± 12.27 | 0.26 ± 0.31 | 87.58 ± 14.87 | 75.27 ± 14.55 |
| **DB-ResNet** | **82.74 ± 10.19** | **0.19 ± 0.21** | **89.35 ± 11.79** | **79.64 ± 13.34** |

To allow a visual comparison of different approaches, the segmentation results are given in Fig. 5. We demonstrated six representative nodules for visual comparison from the LIDC testing set. Notations L1 to L6 shown in Fig. 5 correspond to calcific nodule, juxtapleural nodule, ground-glass opacity nodule, cavitary nodule, isolated nodule, and small nodule less than 6 mm in diameter, respectively. With the visual comparison, it can be seen that the overall performance of the FCN-UNet and MV-CNN methods is slightly inferior to other methods, especially for cavitary nodules and GGO nodules. For isolated nodules, MC-CNN and MCROI-CNN methods performed slightly worse. MCROI-CNN and Cascaded-CNN methods are slightly less effective for juxtapleural nodules. For central calcified nodules, the segmentation results of the MV-DCNN method are incomplete. For small nodules and cavitary nodules, CF-CNN and Cascaded-CNN methods are less adaptable. In contrast, DB-ResNet is still robust when it segments these nodules. This comparison illustrate its significant feature learning capability.

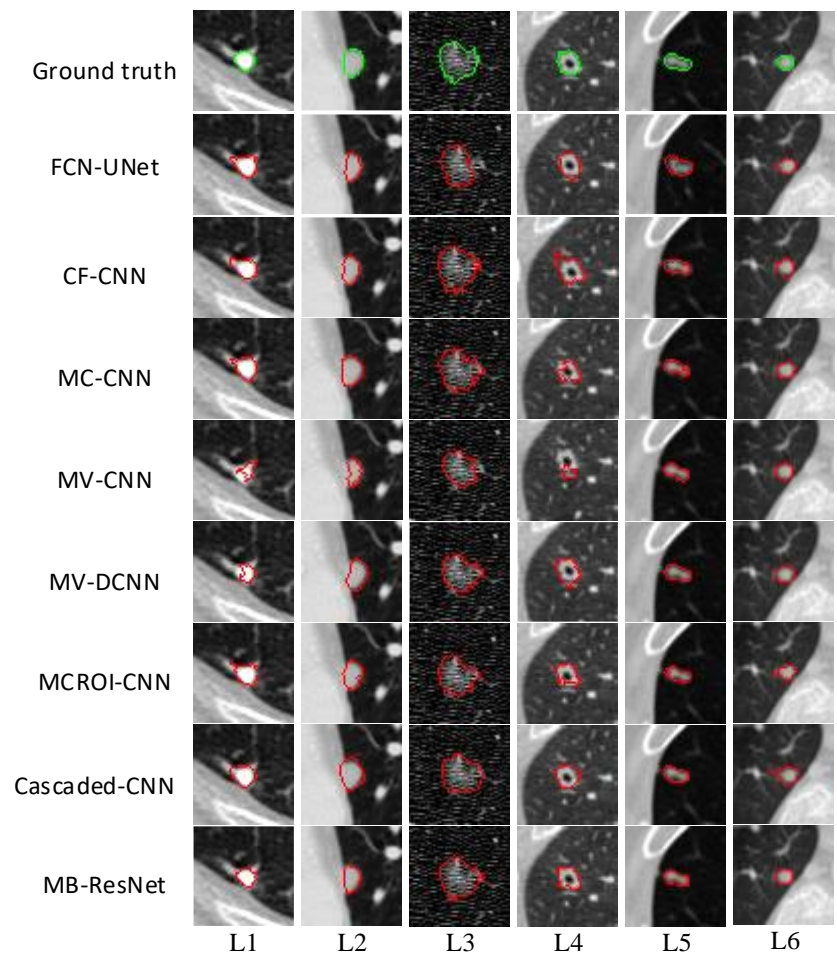

Fig. 5. A visual comparison of the segmentation results. From top to bottom: the ground truth of nodule, segmentation result of CF-CNN, MC-CNN, MV-CNN, MV-DCNN, MCROI-CNN, Cascaded-CNN, and DB-ResNet. Notations L1 to L6 are nodules of different types from the LIDC testing set.

Fig. 6 further shows multiple segmented slices of juxtapleural nodules and small nodules from the LIDC testing set with the application of the DB-ResNet. This comparison indicates that the segmentation results of the DB-ResNet have a large overlap with the ground truth contours.

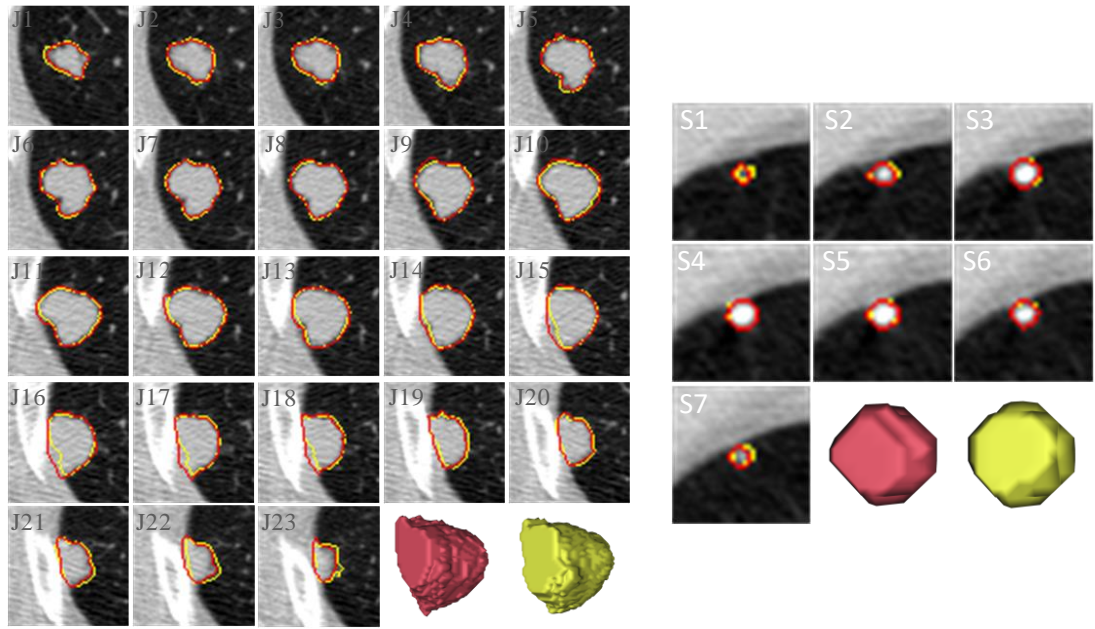

Fig.6. Segmentation results of DB-ResNet on juxtapleural nodule (J1-L23) and small nodule with a diameter of 4.8 mm (S1-S7) from the LIDC testing set. The yellow and red contours represent the segmentation results of DB-ResNet and the ground truth, respectively. The yellow volume data and the red volume data correspond to the 3-D renderings of the DB-ResNet and the ground truth, respectively. The number in the upper left corner of each image represents the slice number where the nodule is located.

# 6. Conclusion

In this study, we proposed a DB-ResNet model for lung nodule segmentation. The model extracts features through dual-branch networks. By comparing with the existing lung nodule segmentation methods, our method showed encouraging accuracy in the lung nodule segmentation task, and the average dice score of 82.74% for the LIDC dataset. Especially, the DB-ResNet model can successfully segment challenging cases such as juxtapleural nodules and small nodules.

In future work, we plan to develop a lung nodule detection algorithm based on the DSSD (deconvolutional single shot detector) network architecture, and then combine it with our method to implement a fully automated segmentation system of the lung nodule.

# Acknowledgements

This work is supported by the National Key R&D Program of China (Grant No.

2017YFC0112804) and the National Natural Science Foundation of China (Grant No. 81671768). The authors acknowledge the National Cancer Institute and the Foundation for the National Institutes of Health and their critical role in the creation of the publicly available LIDC-IDRI Database for this study.